\documentclass[a4paper,11pt]{article}

\usepackage{graphicx}  %%% for including graphics
\usepackage{url}       %%% for including URLs
\usepackage{times}
\usepackage{natbib}
\usepackage[margin=25mm]{geometry}

\usepackage{amssymb}
\usepackage{amsmath}
\usepackage{latexsym}
\usepackage{enumitem}
\usepackage{booktabs}
\usepackage{url}
\usepackage{color} 

\usepackage{arydshln}
\usepackage{mathtools}

\usepackage{soul}

\newtagform{brackets}{[}{]}
\usetagform{brackets}

\newcommand{\csim}{\text{\rm sim}}
\newcommand{\Sim}{\text{\rm Sim}}
\newcommand{\Index}{\mathbb{I}}

\newcommand{\re}[1]{${\tt #1}$}

\title{Semantic Variation in Online Communities of Practice}
\date{}

\author{Marco Del Tredici and Raquel Fern\'andez\\
		Institute for Logic, Language and Computation\\       
       University of Amsterdam\\
       \texttt{\{m.deltredici|raquel.fernandez\}@uva.nl}
}

\begin{document}
\maketitle
\thispagestyle{empty}
\pagestyle{empty}

%==============================================
\begin{abstract}
We introduce a framework for quantifying semantic variation of common words in Communities of Practice and in sets of topic-related communities. We show that while some meaning shifts are shared across related communities, others are community-specific, and therefore independent from the discussed topic. We propose such findings as evidence in favour of sociolinguistic theories of socially-driven semantic variation. Results are evaluated using an independent language modelling task.
Furthermore, we investigate extralinguistic features and show that factors such as prominence and dissemination of words are related to semantic variation.

\end{abstract}

%==============================================

\section{Introduction}
\label{sec:intro}
%Patterns of word usage reflect human activities and interactions. 

%Within the field of Natural Language Processing, 
In computational linguistics and NLP, variation in word meaning has mostly been studied in the abstract, as lists of possible word senses \citep{navigli2009wsd,yarowsky2010wsd}. In contrast, other neighbouring fields such as sociolinguistics and psycholinguistics have emphasised the link between semantic variation and the activities and interactions of speakers. For example, the psychologist Herbert Clark appeals to the notion of `common ground' to characterise patterns of word usage: {\em ``Word knowledge, properly viewed, divides into what I will call {\rm communal lexicons}, by which I mean sets of word conventions in individual communities {\rm [...]} When I meet Ann, she and I must establish as common ground which communities we both belong to simply in order to know what English words we can use with what meaning''} \citep{Clark96}; 
%When I meet Ann, she and I must establish as common ground which communities we both belong to simply in order to know what English words we can use with what meaning''} \citep{Clark96}. 
while the sociolinguist Hasan Ruqaiya argues that {\em ``there is evidence of sociosemantic variation''} which must be taken into account {\em ``unless the concept of meaning is arbitrarily constrained''}  \citep{hasan1989}. 
The distinction between the two approaches is relevant: while the former is based on the idea that, for a given word, a finite list of discrete senses is available, the latter builds on a more dynamic concept, namely that a new meaning can emerge in any interaction among speakers, who use it in order to make communication more effective.

Understanding the intricate ways in which patterns of word use and communities of individuals are related is essential for  characterising the interests and the expressive means of sub-cultures, as well as to develop NLP tools that are effective in the face of variation \citep{hovy:2015:ACL-IJCNLP,YangEisensteinTACL2017}. 
In this paper, we study how word meaning (as captured by distributed vector representations) varies across and within different online communities. We take online communities, such as online discussion forums, to be excellent examples of {\em communities of practice} \citep{wenger2000communities,eckert-mcconnellginet1992}, that is, aggregates of individuals not defined by a location or a population, but rather by social engagement in some common endeavour. Using computational modelling techniques and statistical analyses, we show that community-specific conventional meanings of common word forms (as opposed to jargon) do arise and can be reliably detected, which is consistent with the theoretical standpoint of \cite{Clark96}, among others.

%In contrast to previous work, which has exclusively focused on neologisms or unique jargon --- e.g., {\em `dx'} for `diagnosis' in breast cancer discussion forums \citep{nguyen2011language} or {\em `scrim'} for `practice match' in online gaming \citep{kershaw2016towards} --- we focus on semantic variation of {\em common} words driven by social engagement. Using computational modelling techniques and statistical analyses, we show that community-specific conventional meanings of common word forms do arise and can be reliably detected, which is consistent with the theoretical standpoint of \cite{Clark96}, among others.
%Through such an engagement, {\em ``ways of doing things, ways of talking, beliefs, values, power relations -- in short, {\em practices} -- emerge in the course of their joint activity''} \citep{eckert-mcconnellginet1992}. 

The paper makes the following contributions: We adapt a model for geographically located language introduced by \cite{bamman-dyer-smith:2014} to learn word representations for different online communities of practice. \mbox{We introduce a framework for} quantifying semantic variation and apply it to several Reddit sub-communities engaged in discussing two broad domains, Football and Programming. We evaluate our framework extrinsically against a language modelling task, showing that the semantic shifts we detect on common words are strong enough to affect performance. 
%We then define a range of factors that we hypothesise have an effect on the presence of semantic variation. 

Our results show that distinct meaning conventions arise within communities engaged in discussing a shared domain, but also that the domain itself is not the only determinant of semantic variation: sub-communities concerned with discussing the same general topic may also develop their own conventional meanings for common words, which supports the view that the main factors driving semantic variation are local accommodation effects presumably arising during interaction.

In addition, our findings indicate that, besides frequency-related factors, the level of dissemination of a word among community members plays a key role in understanding the dynamics of meaning variants.

\section{Related Work} 
\label{sec:related}
The present investigation is related to several strands of research in computational sociolinguistics and historical linguistics. Within the former, a substantial amount of work has used NLP techniques to study correlations between linguistic variables and  macro-sociological categories such as age \citep{nguyen2013old}, gender \citep{nguyen2014gender,burger2011discriminating,ciot2013gender}, and other demographic factors \citep{eisenstein2014diffusion}. A related line of research has explored the interplay between language use and social relations among community members. For example, \cite{cassell2005language} and \cite{huffaker2006computational} investigate the correlation between linguistic features and the strength of the relations among users in newborn communities;  \cite{danescu2012echoes} and \cite{noble-fernandez:2015:CMCL} show how variations in linguistic style can provide information about power differences in social groups. Yet other related work has focused on how acceptance into existing communities is mediated by the adoption of community norms \citep{nguyen2011language,tran2016characterizing} and on how the process whereby linguistic innovations become norms can be leveraged to predict the permanence of a user in a community \citep{danescu2013no}.  

Common to all approaches mentioned above is the exploitation of language features to implement predictive models for {\em non-linguistic} features (such as gender, power differences, or community permanence). Less attention has been payed to investigating linguistic variation in its own right. Those approaches that do address this aspect have concentrated almost exclusively on community-specific jargon and slang, i.e., neologisms, unique acronyms and abbreviations --- e.g., {\em `dx'} for `diagnosis' in breast cancer discussion forums \citep{nguyen2011language} or {\em `scrim'} for `practice match' in online gaming \citep{kershaw2016towards}. They have therefore ignored the fact that social interaction among speakers often leads to semantic variation of common word forms: that is, word that {\em ``belong to many communal lexicons, though with very different conventional meanings''} \citep{Clark96}. In the present study we concentrate on precisely this type of semantic variation. 

Our approach takes a synchronic perspective, i.e., we do not look into the temporal dynamics of meaning variation. Nevertheless, in terms of methodology, our work is related to computational historical linguistics. Diachronic meaning change has been studied at different time scales, from a few decades to several centuries. A variety of techniques have been explored:  Latent Semantic Analysis \citep{sagi2011tracing,jatowt2014framework}, topic clustering \citep{wijaya2011understanding} and dynamic topic modelling \citep{frermann2016bayesian}. More recently, word embeddings \citep{mikolov2013efficient} have proved useful for investigating meaning change over time. The most common approach consists in creating independent vector representations for consecutive time spans and then using a transformation matrix to map vectors from one space to another one \citep{kulkarni2015statistically,zhang2015omnia,hamilton2016diachronic}. Similarly to this strand of research, our work leverages the power of word embeddings, but exploits a different approach originally introduced by  \cite{bamman-dyer-smith:2014} to account for geographical variation. As we will explain in detail in Section~\ref{sec:framework}, this approach is an extension of the skip-gram vector model \citep{mikolov2013efficient} that allows us to learn meaning representations per community that build upon shared representations. 

Meaning variation determined by geographical location --- including that of \cite{bamman-dyer-smith:2014} --- has often focused on dialectal varieties in the USA using data from Twitter \citep{eisenstein2010latent,doyle2014mapping,eisenstein2014diffusion}. In contrast to this line of work, as pointed out in the introduction, we are interested in investigating semantic variation in {\em communities of practice} \citep{wenger2000communities,eckert-mcconnellginet1992}: communities defined by social engagement rather than geo-location or other demographic variables. 

\section{Experimental Setup}
\label{sec:exp}
\begin{figure*}[t]
\begin{minipage}{4.5cm}
\includegraphics[height=4.2cm]{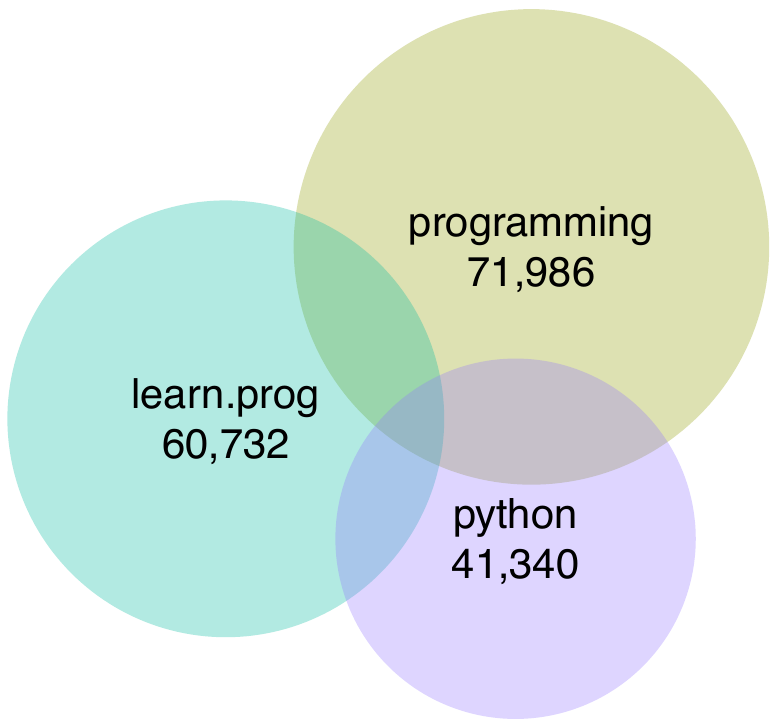}
\end{minipage} \ \ 
\begin{minipage}{4.5cm}
\includegraphics[height=4.2cm]{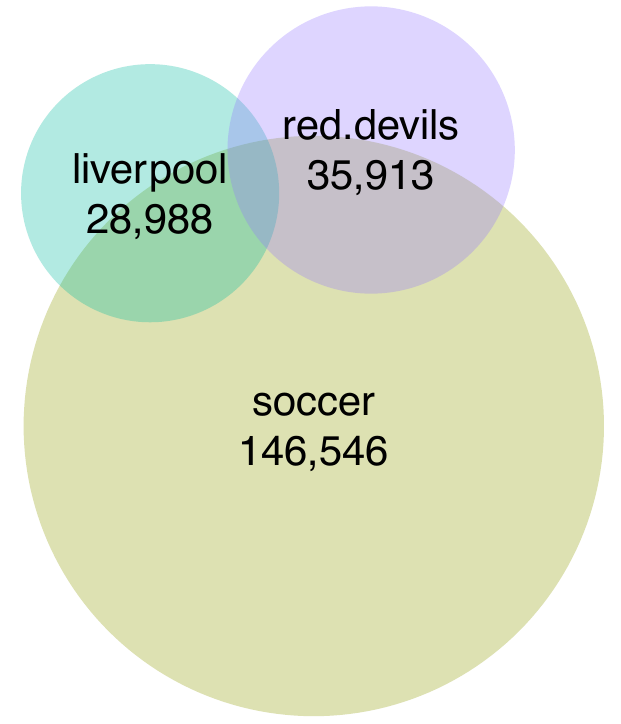}
\end{minipage}
\begin{minipage}{4.5cm}\centering %\small
\begin{tabular}{@{}lcc@{}}
\bf community & \bf years & \bf million tokens\\\hline
\re{programming} & 10 & 21\\
\re{python} & 8 & 18\\
\re{learn.prog} & 7 & 21\\\hdashline
\re{soccer} & 8 & 65\\
\re{liverpool} & 8 & 55\\
\re{red.devils} & 6 & 66\\\hdashline
global & -- & 50\\\hline
\end{tabular}
\end{minipage}
\caption{Left: total number of members and their overlap in the Programming and Football supra-communities. Right: main statistics (time span and number of word token) in each community dataset.}\label{fig:reddit}
\end{figure*}

Online communities offer an unprecedented opportunity to study linguistic variation and its dynamics. For our investigation of semantic variation, we collected data from Reddit, a large on-line community which includes approximately 1 million sub-communities called `subreddits'.\footnote{\url{https://www.reddit.com}} A subreddit is essentially a discussion forum where individuals with a shared interest on a topic or activity interact: once a user has subscribed to a subreddit, she can post any kind of content (text, links, pictures), reply to existing posts as well as `upvote' or `downvote' them. Subreddits can therefore be considered {\em communities of practice} in the sense of \citep{eckert-mcconnellginet1992}.

We collected data from 6 different subreddits: half of them (\re{programming}, \re{learn.programming} and \re{python})\footnote{The actual names of the latter two subreddits are \re{learnprogramming} and \re{Python}; we have slightly modified the names for clarity and simplicity.} are concerned with the domain of computer programming, while the other half (\re{soccer}, \re{liverpool}, and \re{red.devils}) are related to the domain of sports, in particular football.\footnote{The subreddit \re{liverpool} (actual name \re{LiverpoolFC}) consists of fans of Liverpool Football Club, while \re{red.devils} (actual name \re{reddevils}) groups fans of the Manchester United Football Club.} 
%\raq{we may need to mention that soccer and programming are ``more general''...}
%\mdt{I think this is related to the the fact that we don't see community shift for these two communities}
We refer to the subreddits as {\em communities} and to the two groups of subreddits related by a common theme, Programming and Football (with a capital), as {\em supra-communities} or domains. It is important to note that Reddit does not have a hierarchical structure whereby subreddits are classified into groups.  We base this grouping on the common theme and on shared membership. The communities \re{programming} and \re{soccer} have a somewhat special status as they are more general in terms of topic and larger in terms of number of members.
Figure~\ref{fig:reddit} shows the total number of members in each community and the pattern of shared membership within a supra-community. Over 12\% and 15\% of members within Programming and Football, respectively, belong to at least two communities in the respective domain. The communities within a domain are thus substantially interconnected. In contrast, the Programming and Football supra-communities share less than 2\% of users. 
%Finally, we note how and \re{soccer} is, in the Football domain, the more general community. Again, this can be easily seen by looking at the topics of the other communities in the same domain, and at the data about common users:  46\% of \re{liverpool} users and 47\% of \re{red.devils} users are also members of \re{soccer}. What said for \re{soccer} is partially valid also for \re{programming}: while 24\% of the users in \re{python} are also in \re{programming}, \re{learn.programming} shares with \re{programming} only 12\% of the users. 

For each community, we crawled the contents created by all members during its whole lifespan (between 6 and 10 years). In the present study, we do not make use of the longitudinal character of the corpus, i.e., we abstract away from the temporal aspect and consider each of the community datasets synchronically as a whole.\footnote{The temporal information is a valuable feature of the corpus, which we plan to exploit in future work -- see Section~\ref{sec:concl}.} Since the resulting datasets had different sizes in terms of number of tokens, we randomly subsampled some of them (those crawled from \re{soccer}, \re{programming}, and \re{learn.programming}) in order to make them comparable in size to the other communities within the same domain. The table in Figure~\ref{fig:reddit} summarises the main statistics for each community.

Finally, in order to obtain a sample of community-independent linguistic practices, we created an additional dataset by randomly crawling posts and comments exchanged within any of the existing subreddits during January 2017. We refer to this dataset as the {\em global} community. The global community includes 50 million tokens from hundreds of thousands of different subreddits, contributed by more than 445k different users. Less than 1\% of these users are members of the Programming and Football supra-communities. We consider the linguistic practices present in this dataset as a proxy for general language use. 

This experimental setup allows us to investigate different types of semantic variation taking place at different levels: (1) meaning variants deviating from the general language and shared by communities concerned with a common domain, and (2) meaning variants specific to a community and differing both from general language and from other communities within the same domain. In the next section, we define a framework for capturing these two types of semantic shift in a precise, quantitative manner.

\section{Framework}
\label{sec:framework}
We describe the vector-space model we use to learn word representations for online communities of practice and then introduce two indices to measure semantic variation. 
%The framework is defined in a general fashion and is then applied to the particular corpora described in the previous section.

\subsection{Vector space model}
\label{sec:vectorspacemodel}
Let $C$ be a set of {\em communities of practice} and let $g$ denote the {\em global community}, reflecting general (community-independent) language use. We use subsets such as $D\subset C$ to denote sets of communities related by a certain \emph{domain} (Programming and Football in the experimental setup we use here). 

We adapt the model introduced by \cite{bamman-dyer-smith:2014} for geo-located language, which in turn is an extension of the skip-gram model by \cite{mikolov2013efficient}. The model relies on a set of contextual variables---geographical locations in the case of \cite{bamman-dyer-smith:2014} and \cite{kulkarni2016freshman}, and online communities of practice in our setup. Instead of using a single embedding matrix containing a single real-valued vector for every word in the vocabulary, several matrices are defined: a main matrix $W$, which is learned by considering all occurrences of each target word in the entire corpus, and one matrix $W_c$ per community (including one matrix $W_g$ for the global community). During training, given an input word $w$ used in a message exchanged within community $c$, the hidden layer is calculated as the sum $h = w^{\top}W + w^{\top}W_c$. Back-propagation via stochastic gradient descent then updates both embedding matrices. 

This joint parametrisation has several desirable properties: the model learns different embeddings for the same word (one per community: $w_{c1}, w_{c2},\ldots, w_g$) that are part of the same vector space and therefore can be compared to each other. Furthermore, the word representations share information across communities (via the main matrix $W$), which, intuitively, operates as a regularizer, thus capturing the intuition that the use of a word in a given community is not radically different from its use in other settings but rather a modulation of conventions built upon general shared common knowledge \citep{Clark96}.

We tokenise the datasets described in Section~\ref{sec:exp} (no further preprocessing is applied), and create two independent vector space models for the Programming and Football supra-communities, respectively. We consider only those words that appear at least 100 times in each community dataset and learn word embeddings with 200 dimensions using L2 regularisation. The global community dataset is used in both models. 

%================================
\subsection{Measures of semantic variation}
\label{sec:measures}

The model described above allows us to derive word embeddings $w_c$, $w_g$ for each word $w$ and community $c$ in a domain $D$, encoding how $w$ is used within that community and in the global community $g$, respectively. 
 
For any two vectors $v,v'\in\mathbb{R}^k$, let $\csim(v,v')$ denote their \emph{cosine similarity}. Given two sets of communities $A,B\subseteq C\cup\{g\}$, we use $\Sim^w_{A,B}$ to refer to the following multiset of similarity values %\footnote{Recall that, unlike a set, a multiset allows multiple instances of the same elements. This is appropriate in our case since cosine similarity values for different pairs of words may be identical.} 
for word $w$:\footnote{In practice, in case $A=B$, we only compute one cosine similarity value for every unordered pair rather than for every ordered pair. Observe that this does not affect either the mean or the standard deviation of the multiset.} 
%\[\begin{array}{r@{\ }c@{\ }r}
%\Sim^w_{A,B} & = & \{ \csim(w_a,w_b) \mid (a,b)\in A\times B\ \ \ \ \\ 
%	&& \text{with}\ a\not=b\}
%\end{array}\]

\[\begin{array}{r@{\ }c@{\ }r}
\Sim^w_{A,B} & = & \{ \csim(w_a,w_b) \mid (a,b)\in A\times B \ \text{with}\ a\not=b\}
\end{array}\]

\noindent
Let $S$ and $S'$ be two such multisets of similarity values. To measure the extent to which these values are higher in $S$ than in $S'$, we use the following index, where $\mu$ and $\sigma$ are the mean and the standard deviation, respectively:
\begin{eqnarray*}
\Index(S,S') & = & [\mu(S)-\sigma(S)] - [\mu(S')+\sigma(S')]
\end{eqnarray*}

\noindent 
We can now use this generic index to construct several specific indices to quantify different types of semantic variation.

%==== dsi =======
 \paragraph{Variation at domain level:} We consider that a word $w$ exhibits a domain-specific semantic shift if its meaning is relatively constant across communities with a common domain, while being distinct from its use in the global community. The \emph{domain shift index} ${\bf dsi}^w(D)$ captures exactly this, for a given domain $D\subset C$ and word~$w$:
\begin{eqnarray*}
{\bf dsi}^w(D) & = & \Index(\Sim^w_{D,D},\Sim^w_{D,\{g\}})
\end{eqnarray*}

\noindent
For words with positive {\bf dsi} values, the higher the index, the more pronounced their semantic shift across a domain with respect to the language use of the global community.

% ==== csi ====
  \paragraph{Variation at community level:} We now want to quantify the degree to which a given word exhibits a semantic shift specific to a community, i.e., not shared by other communities concerned with the same domain $D$. This type of semantic variation is particularly interesting because, when present, it arguably shows that meaning variants can arise in a community independently from the topic discussed. 
  
In particular, we focus on capturing scenarios where the meaning of a word $w$ in a community $c \in D$ has drifted away from its general use in $g$, while in other domain-related communities the meaning remains closer to that observed in the global community. This is what the {\em community shift index} below captures, where $D\setminus\{c\}$ denotes the set of communities in domain $D$ except for $c$:  
\begin{eqnarray*}
{\bf csi}_{c}^w(D) & = & \Index(\Sim^w_{D\setminus\{c\},\{g\}},\Sim^w_{\{c\},\{g\}})
\end{eqnarray*}  
  
\noindent
Again, for words with positive ${\bf csi}_{c}^w(D)$ values, the higher the index, the stronger the shift in $c$ relative to other domain-related communities.
  
% ===

Using the community-specific word embeddings learned with our vector space model, we compute ${\bf dsi}^w(D)$ and ${\bf csi}_{c}^w(D)$ values for all words per domain and community, respectively.

\begin{figure*}[t]\centering
\includegraphics[width=10cm]{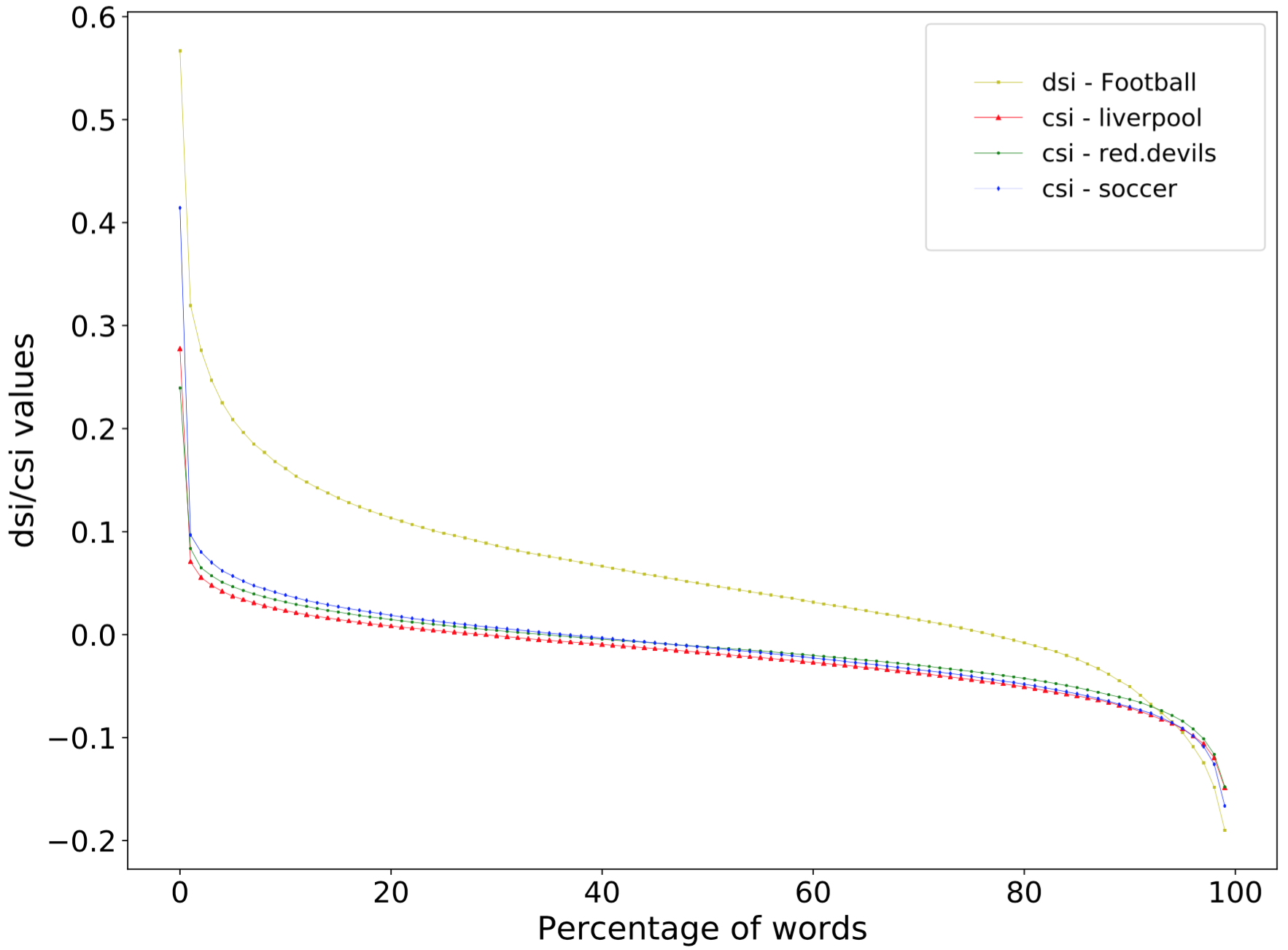}
\caption{The {\bf dsi} and {\bf csi} values (on the y-axis) for the Football domain and the communities which belong to it. On the x-axis the number of words (in percentage): since we consider only words in the shared vocabulary (see Section~\ref{sec:exp}) the total amount of words is the same for all the communities and for the domain.}
\label{fig:values_distribution}
\end{figure*}

Figure~\ref{fig:values_distribution} shows the distribution of {\bf dsi} values for the Football domain and the {\bf csi} values of the communities belonging to the domain.\footnote{Similar results are found for the Programming domain and its communities.} All the distributions present a common pattern: few words undergo a strong semantic shift in the domain / community (left tail of the graph), while the majority of the words present a small or null shift, corresponding to {\bf dsi} / {\bf csi} values included in the range between 0 and 0.2. Note that on average {\bf dsi} values are larger than {\bf csi} ones because, intuitively, the {\bf dsi} captures the shift in the domain vocabulary compared directly to the global community, while {\bf csi} represents the more subtle shifts within communities belonging to the same domain. The right tail of negative values has different interpretations for the domain and the communities. Negative values of {\bf dsi} are assigned to the same words that have high {\bf csi} values, i.e. words that show a strong shift in just one of the communities part of the general domain. Finally, for each community, negative {\bf csi} values are assigned to words that undergo strong semantic shift in \textit{another} community of the same domain.

%==============================================
\section{Evaluation}

In order to verify whether the measures proposed in the previous section capture semantic variation that is noticeable beyond cosine distances in semantic space, we evaluate them using an independent language modelling task.  

\subsection{Method}

We implement a neural language model (NLM) using an existing encoder-decoder LSTM\footnote{\url{https://github.com/pytorch/examples/tree/master/word_language_model}} with 2 layers of size 200. We randomly split the dataset of each community into training (70\%), validation (15\%), and test (15\%) sets and train one NLM per community using the word embeddings previously learned for that community with the vector space model described in Section~\ref{sec:vectorspacemodel}. We train the models for 40 epochs, using Adam  estimation \citep{kingma2014adam} for parameter update and dropout for regularisation. The same procedure is also carried out for the global community. All the community language models reached an average test perplexity between 45 and 67 on the task of predicting the upcoming word given the preceding word (window size = 1) --- a performance in line with the state of the art, (e.g., \cite{zaremba2014recurrent}).

%For the evaluation of our domain and community shift indices ${\bf dsi}^w(D)$ and ${\bf csi}^w_c(D)$, 
For each domain $D$, we define two sets of target words: a \re{shift} set containing the top 10 words with the highest ${\bf dsi}^w(D)$ values, and a \re{no.shift} set containing the 10 bottom words with the lowest positive ${\bf dsi}^w(D)$ values. 
We do the same per community $c$: the \re{shift} set includes the ten words with the highest ${\bf csi}^w_c(D)$, while the \re{no.shift} set includes the ten words with the lowest  ${\bf csi}^w_c(D)$ per communtiy.

At test time, given a set of target words, we compute the average perplexity for each target word $w$ on predicting $w+1$ with the original $w$ embeddings used for training (${\rm ppl}^w_{train}$) and with \textit{alternative} embeddings for $w$ learned from another community (${\rm ppl}^w_{alt}$). We then measure change in performance as relative perplexity increase:
\begin{eqnarray*}
{\rm ppl}^w_{change} & = & \frac{{\rm ppl}^w_{alt} - {\rm ppl}^w_{train}}{{\rm ppl}^w_{train}}
\end{eqnarray*}
The rationale behind this method is the following: Regarding domain variation, we hypothesise that for \re{shift} words the increase in perplexity of the NLM of a given community will be significantly higher when testing on alternative embeddings belonging to the general community than on alternative embeddings belonging to another domain-related community. Regarding community-specific variation, we hypothesise that, when leveraging the NML of the global community, using embeddings of \re{shift} words in community $c$ as alternative embeddings will yield significantly higher perplexity than using alternative embeddings from other communities within the same domain.\footnote{Recall that $\textbf{csi}^w_c(D)$ is meant to capture a meaning variant of $w$ in $c$ that has drifted away from $w$'s use in the global community more than in other domain-related communities.} In all cases, we expect that for \re{no.shift} words (i.e., words for which there is no semantic variation according to our indices) the change in perplexity with different embeddings will be negligible. 

We evaluate these hypotheses by calculating ${\rm ppl}^w_{change}$ values for  \re{shift} and \re{no.shift} words and checking for significance with Wilcoxon signed-rank test.

\begin{table} \centering
\begin{tabular}{@{}l@{}cc@{\ }c@{\ }|cc@{}}
                  & \multicolumn{3}{c|}{\re{shift}} & \multicolumn{2}{c}{\re{no.shift}}\\\toprule
                  \bf domain & $c\!\rightarrow\!{D\!\setminus\!\!\{c\}}$ & $c\!\rightarrow\!g$ & & $c\!\rightarrow\!{D\!\setminus\!\!\{c\}}$ & $c\!\rightarrow\!g$   \\\hline
Programming & 6.04 & 64.9 &  (*)& 5.77 & 9.02 \\
Football          & 2.40 & 40.47 & ** & -0.78 & -1.04\\\midrule
\bf community &  $g\!\rightarrow\!{D\!\setminus\!\!\{c\}}$ &$g\!\rightarrow\!c$&  & $g\!\rightarrow\!{D\!\setminus\!\!\{c\}}$ & $g\!\rightarrow\!c$\\\hline
\re{programming} & 0.24 &  3.87 &     & 4.92 & 10.05 \\
\re{python}           & 6.73 & 26.83 & ** &  0.68 & 8.83 \\
\re{learn.prog}      &11.85 & 56.77 & * &  9.57 & 13.28 \\\hdashline
\re{soccer}            & 11.32 & 8.92  &     & 13.33 & 11.93\\
\re{liverpool}         &  2.45 & 17.70 & ** &  3.84 & 5.31 \\
\re{red.devils}       & 4.98 & 55.84 &  ** & 2.98 & 5.60  \\\bottomrule
\end{tabular}
\caption{Perplexity increase medians in each setting with significance level of Wilcoxon signed-rank test (***$p\!<\!.001$, **$p\!<\!.01$, *$p\!<\!.05$).\vspace*{-5pt}}
\label{tab:lm}
\end{table}

\begin{figure*}\centering
\begin{minipage}{5.2cm}
\fbox{\includegraphics[width=5.05cm]{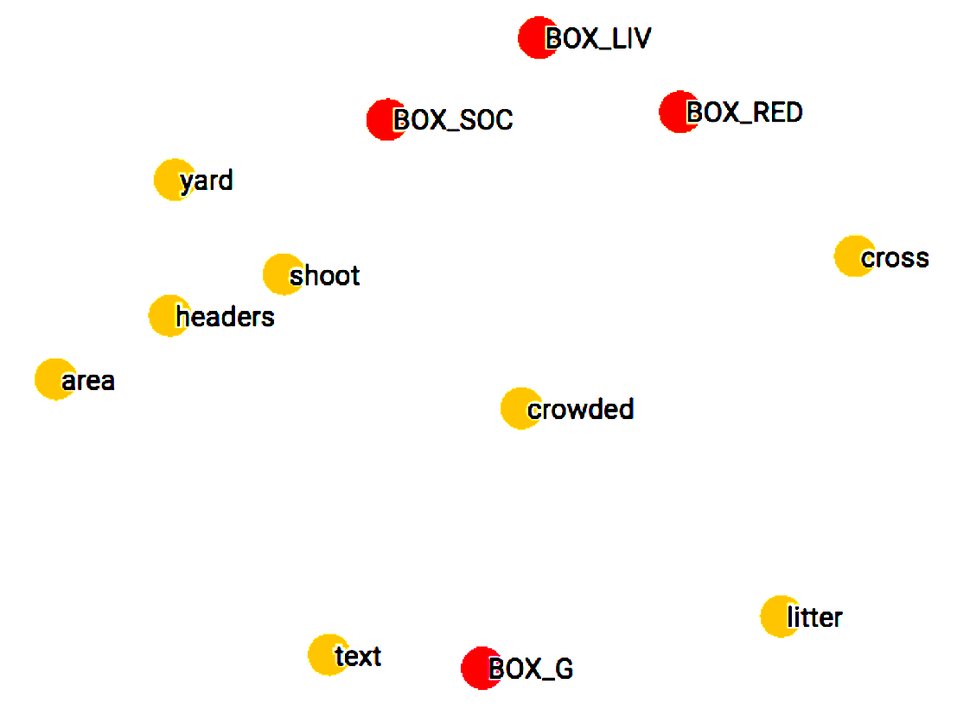}}
\end{minipage} 
\begin{minipage}{5.2cm}
\fbox{\includegraphics[width=5.05cm]{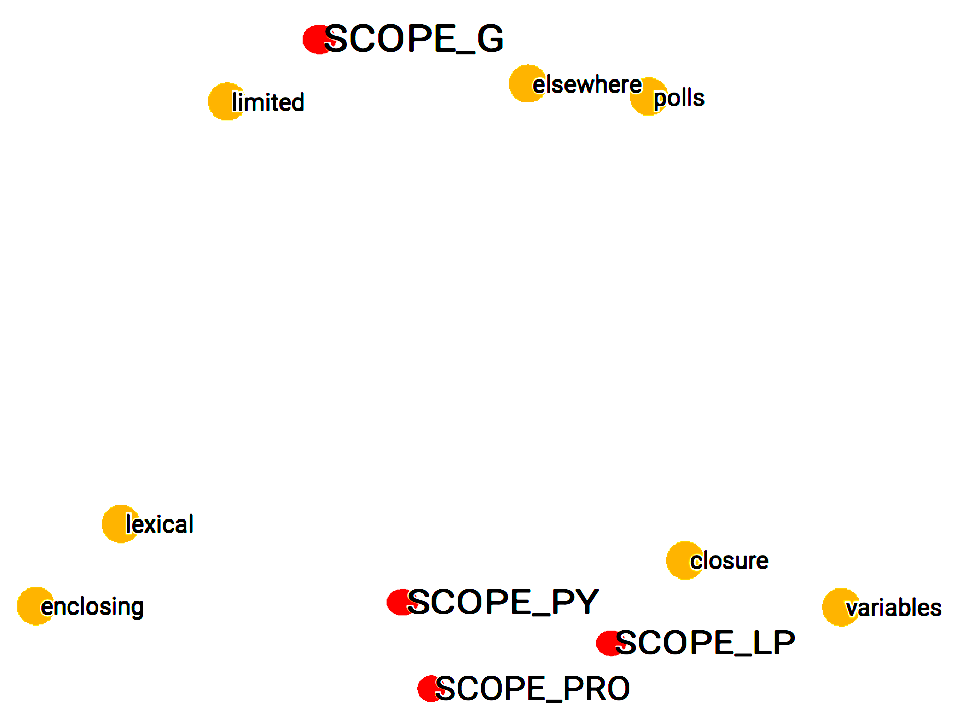}}
\end{minipage} 
\begin{minipage}{5.2cm}
\fbox{\includegraphics[width=5.05cm]{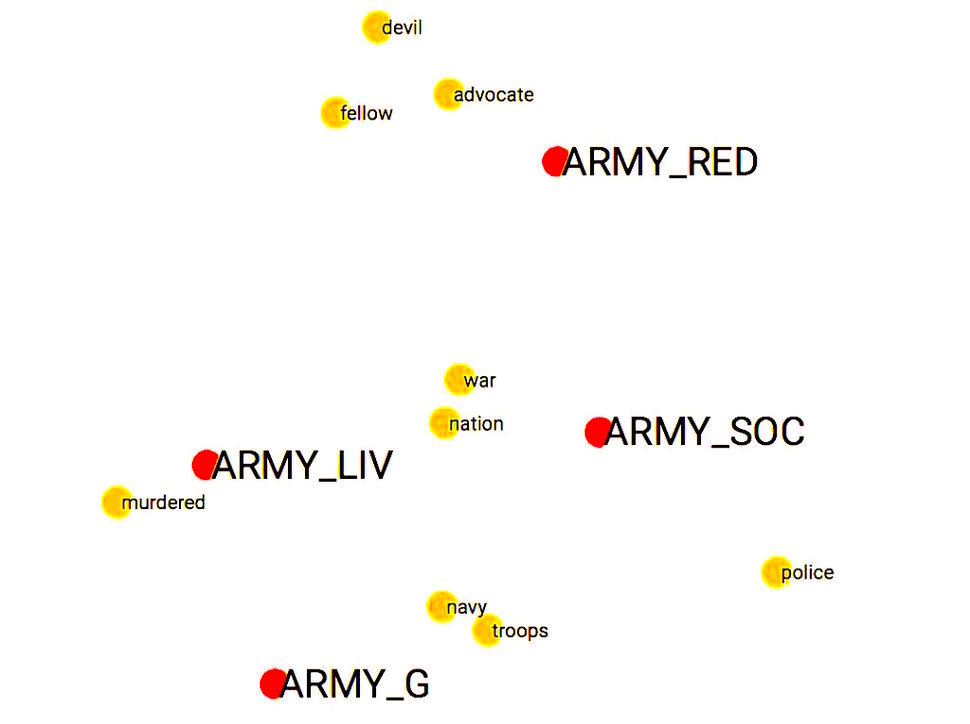}}\end{minipage} 
\caption{Two-dimensional representation in semantic space of meaning variants for words `box' (high  \textbf{dsi}(Football)), `scope' (high \textbf{dsi}(Programming)), and  `army' (high $\textbf{csi}_{\tt red.devils}$(Football)).}\label{fig:semantic_spaces}
\end{figure*} 

%army 
%csi = 0.953 (3rd)
%fre=  0.129 (medium part of the list)
%
%box 
%dsi = 0.933 (6th)
%fre = 0.524 (note, despite the low value, it is still at the top of the list)
%
%scope = 
%dsi = 0.851 (4th)
%fre = 0.365 (note, despite the low value, it is still at the top of the list)
%

\subsection{Results}
\label{sec:lm-results}

Table~\ref{tab:lm} shows an overview of the results. For conciseness, we only show the median ${\rm ppl}^w_{change}$ values.%\footnote{All values are available in the supplementary material submitted with the paper.} 
Regarding domain variation, as predicted, for words with low ${\bf dsi}^w(D)$ values (\re{no.shift}), we never observe a significant difference in perplexity when different embeddings are used. In contrast, for words with high ${\bf dsi}^w(D)$ values (\re{shift})  the increase in perplexity is always significantly higher when the original embeddings of a community are substituted with those of the general language ($c\!\rightarrow\!g$), while perplexity remains reasonably stable when the alternative embeddings come from another domain-related community ($c\!\rightarrow\!{D\!\setminus\!\!\{c\}}$). This holds for both domains, Football and Programming, with the exception of the \re{programming} community,  for which there is no significant difference in perplexity when \re{learn.programming} and general language embeddings are used ---  indicated by (*) in Table~\ref{tab:lm}. 

As for community-specific variation,  again we never observe a significant difference in perplexity for words with low ${\bf csi}_{c}^w(D)$ values  (\re{no.shift}). For words with high ${\bf csi}_{c}^w(D)$ values  (\re{shift}), our hypothesis is confirmed for the more specific communities  \re{liverpool}, \re{red.devils}, \re{python} and \re{learn.programming}: there is a significant increase in perplexity when the embeddings from these communities are used with the global NLM ($g\!\rightarrow\!c$), which is in line with the presence of a community specific semantic variant within the domain. This is not confirmed for the more general communities \re{programming} and \re{soccer}. This latter negative result is in fact intuitive: it is unlikely that these more general and larger communities will exhibit meaning variants that are further away from general language use than the more specific, smaller communities. 

%\begin{figure}\centering
%\fbox{\includegraphics[width=6cm]{figures/scope.jpg}}\\
%\fbox{\includegraphics[width=6cm]{figures/rsz_army.jpg}}
%\caption{Semantic variants of word `scope' (high  \textbf{dsi}(Programming)) and word  `army' (high  $\textbf{csi}_{\tt red.devils}$(Football)).}
%\end{figure}

Figure 2 shows some examples of meaning shift captured by our indexes. The words `box'  and `scope' are among the ten words with the highest \textbf{dsi} for Football and Programming, respectively. As a consequence, the domain-related variants are closely located, while the variant of the general community is farther away in  semantic space. Difference in meaning is also evident from the nearest neighbours.\footnote{In the domain of Football, `box' has come to mean the penalty area, which is associated with game actions such as `cross', `shoot' and `headers'.\label{ftn:box}} 
The word `army' has high {\bf csi} in the \re{red.devils} community. In the other domain-related communities, the word has meaning variants that are closer to its use in the general community. In the \re{red.devils} community, however, `army' is conventionally used to denote the Manchester United fans (e.g., `we need all types of supporters to make the red army'), as evidenced by its closest neighbours. 

\section{Factors Influencing Semantic Variation}
\label{sec:factors}
Having confirmed that the semantic shift indices proposed in Section~\ref{sec:measures} capture  variation that is noticeable in an external language modelling task, we now turn to analysing the factors that may be related to the presence of such variation.  

\subsection{Features}
\label{sec:factors-features}

We consider four features capturing different properties of word {\em forms} and investigate their effect on meaning variation:

% all features in the range [0,1] ?

\paragraph{Frequency.} 
It is known that more frequent words have a tendency to be more polysemous \citep{Zipf49}, are more semantically stable over time \citep{hamilton2016diachronic}, and evolve at slower rates across languages \citep{pagel2007frequency}. Word frequency may therefore play a role in semantic variation across communities of practice. We compute word frequency as the log-scaled relative frequency of a word in a given community: 
\begin{eqnarray*}
{\rm Freq}(w,c) & = &  \log_{10}(N^w_c / N_c)
\end{eqnarray*}
where $N_c$ is the total number of words in the sample dataset of community $c$ and $N^w_c$ the number of occurrences of word $w$ in that sample. Frequency in a domain ${\rm Freq}(w,D)$ is calculated equivalently.

\paragraph{Prominence.} 
Many measures have been proposed to weight the prominence of a word in a language sample, including TF-IDF. Our choice here is inspired by literature on terminology extraction \citep{velardi2007termextractor}. We compute the prominence of $w$ as its frequency in a community ($N^w_c$) relative to its frequency in a domain, or as its frequency in a domain ($N^w_D$) relative to its frequency in general language use ($N^w_g$):
\begin{eqnarray*}
{\rm Pro}(w,c) & = & N^w_c / (N^w_c + N^w_{D\setminus\{c\}})\\
{\rm Pro}(w,D) & = & N^w_D / (N^w_D + N^w_g)
\end{eqnarray*}
Community-specific jargon or slang words will typically have very high prominence. In contrast, we hypothesise that common words exhibiting semantic variation as a result of community conventions --- which are our focus here --- are likely to {\em not} be singled out by very high  prominence values.  Nevertheless, their level of prominence may still be a determiner of variation.

%Given our hypothesis that semantic variation is not restricted to jargon (technical terms) but may affect common words as a result of the dynamics of interaction, we expect to find words that undergo a semantic shift even if they do not exhibit high prominence values. 

%\raq{This is indeed related to the issue of common words vs technical words/jargon. We will need to refine this as we make this aspect clearer in earlier sections of the paper.}
%\mdt{I have added parts related to this in 4.2 and 5. I think that the way you state out hypothesis here is fine}
%MARCO: As you said several times, frequency and relevance are related, so probably the hypotheses for the two of them will be the same. I don't have a clear idea now: probably, since we want to consider common words, exactly the fact they have similar DR and freq in general lanaguage and domain/community make them 'common', and therefore I don't expect domain and common shift to go up as freq and dr go up. Yes, now that I think about it, I think that exactly these two features can be the solution for identifying 'common words'. Let me know what you think.

\paragraph{Specificity.} 
Besides frequency-related aspects, we also want to capture the extent to which a given word $w$ appears in a restricted set of contexts. We approximate this by computing the collocational score of every bigram containing $w$ and then scoring them using log-likelihood ratio as association measure \citep{dunning1993accurate,manning1999foundations}.\footnote{We used the NLTK implementation described at \url{http://www.nltk.org/howto/collocations.html}} We take the value of the highest ranked bigram as a proxy for the contextual specificity of $w$ in community $c$ (${\rm Spe}(w,c)$) or domain $D$ (${\rm Spe}(w,D)$). The feature values are normalised to obtain scores in the range $[0,1]$.

%In line with our observations on semantic narrowing as one of the processes leading to semantic variation, we hypothesise that words that exhibit a semantic shift are likely to have higher contextual specificity.
%\raq{so far, there are no observations on semantic narrowing in the qualitative analysis}

%\textbf{contextual specificity}
%\underline{domains} I would expect a positive correlation between this feature and domain shift. Strong collocation can be seen as the step coming before semantic narrowing: a word that in the general language is used in many different contexts, tends to be used only in specific contexts in a domain, and this is highlighted by an high collocation index. The following step is the one in which there is no need to specify the context anymore, and the narrowing process is complete. Example of this scenario are 'yellow' (highly collocated with 'card', but also occuring alone with the narrowed meaning) and ‘window’ ('market window' - 'window').
%
%\underline{communities} I expect the positive correlation for communities. An example is 'theatre' in reddevils: 'theatre of dreams' is the way they call their stadium. This occurs of course very often (strong collocation). But you can find 'theatre' alone as well, with its narrowed meaning of (Manchester) 'stadium'

\paragraph{Dissemination.} Finally, we consider the range of individuals using a given word. A priori, words with the same frequency, prominence, or contextual specificity may differ in their level of social dissemination, i.e., in the proportion of community members using them. We compute a word's dissemination within a community $c$ as follows: 
\begin{eqnarray*}
{\rm Dis}(w,c) & =  & (U^w_c / U_c) \times (1 - {\rm RelFreq}(w,c))
\end{eqnarray*}
where $U^w_c$ is the number of community members who use word $w$ and $U_c$ the total number of members in community $c$. Since words with very high frequencies (such as function words) will be used across the board, we weight the ratio $U^w_c / U_c$ by the inverse of $w$'s relative frequency. Dissemination in a domain ${\rm Dis}(w,D)$ is calculated equivalently.

Word dissemination has been shown to be predictive of changes in word frequency over time \citep{altman-etal:plos-one:2011}. Here we investigate whether it is a determiner of semantic variation. 
%Given our emphasis on social interaction, we expect it to be a key factor: for a shift to affect several domain-related communities, the word must be sufficiently disseminated among them; for a shift to be exclusive of a single sub-community the word can't be highly disseminated if the community shares a substantial number of members with other communities (only if it is isolated from other communities could we expect high dissemination).

%\textbf{dissemination} 
%
%\underline{domains}I expect a positive correlation: if a word undergoes a shift it means that is important in that domain and, in turns, this means that is used by a large number of users.
%
%\underline{communities} I expect positive correlation here as well, for the same reason above. I can also think to an example that I saw in the data: the word 'believer' in LiverpoolFC has high community shift index, and is particularly interesting because until 2016 it was not spread at all; then it was used during a famous interview from the team coach in 2016, and from the moment on it was used by an increasing number of users of the subreddit (i.e. more dissemination). Now that I think about it, probably my hypothesis is somehow biased by the fact that I saw phenomenon of this kind in the data. 
%

\subsection{Results}
\label{sec:factors-results}

\begin{table} \centering
\begin{tabular}{@{}l@{\; \  }l@{\; \ }l@{\; \ }l@{\; \ }l@{}}
& ${\rm Freq}$ & ${\rm Pro}$ & ${\rm Spe}$ & ${\rm Dis}$\\\toprule
Programming & 0.96 *** & 0.72 *** & 0.46 *** & 0.37 ***\\
Football& 1.32 *** & 0.81 *** & 0.52 *** & 0.63 *** \\\midrule
%\re{programming} & -- & -- & -- & -- \\
\re{python} & 0.12  & 1.0 \ \  *** & 0.09 \;  &  0.21 ***\\
\re{learn.prog} & 0.02 & 0.63 *** & 0.32 *  & 0.23 **\\\hdashline
%\re{soccer} & -- & -- & -- & -- \\
\re{liverpool} & 0.26  ** & 0.47 ** & 0.18  & 0.38 *** \\
\re{red.devils} & 0.16 &  0.42 ** & 0.16  & 0.20 *  \\\bottomrule
\end{tabular}
\caption{Effect size (Cohen's $d$) and unpaired two-sample $t$-test significance level (***$p\!<\!.001$, **$p\!<\!.01$, *$p\!<\!.05$) for each feature.}\label{tab:factors}
\end{table}

To investigate the role of the features introduced above, we test whether their values are significantly different in words that exhibit a strong semantic shift (words with  {\bf dsi} / {\bf csi} values equal or larger than 2 standard deviations above the mean within a domain or community) and words with no semantic variation (with index values lower than one standard deviation above the mean).

%according to our indices and words that do not. We extend the approach of Section~\ref{sec:eval}: instead of taking only the 10 top and bottom words in the positive range of {\bf dsi} / {\bf csi}, we consider all words with {\bf dsi} / {\bf csi} values above 2 standard deviations above the mean to exhibit a strong semantic shift, and those words whose indices are lower than one standard deviation above the mean to exhibit no shift. 

At the domain level, we find very robust patterns for all features: the words that have undergone a strong domain shift have significantly higher frequency, prominence, contextual specificity, and social dissemination  in each respective domain, Programming and Football. Table~\ref{tab:factors} shows the significance level of a unpaired two-sample $t$-test and the effect size for each feature.

At the community level, since the shifts for the \re{programming} and \re{soccer} communities were not validated in our extrinsic evaluation (Table~\ref{tab:lm}), we do not consider these communities here. For the other 4 communities, we find a systematic pattern: words that exhibit a semantic shift particular to a community are significantly {\em more} prominent in that community than in other domain-related communities, and {\em less} disseminated within that community than words that do not exhibit a shift. The significance of frequency and specificity vary per community. A summary is given in Table~\ref{tab:factors}.

\subsection{Qualitative analysis}
\label{sec:factors-analysis}

% Pro sig more in shift but not jargon. 
 
As hypothesised, words with high \textbf{dsi} /  \textbf{csi}  values have significantly higher levels of prominence in the respective domain or community, but lower levels than jargon. For example, `box' and `believers', which have high \textbf{dsi} in Football and high  \textbf{csi}  in \re{liverpool}, respectively, have prominence values of 0.7, in contrast to jargon terms such as `hat-trick' (Pro=1 in Football) and `bitwise' (Pro=1 in Programming), which are not singled out by our semantic shift indices. This confirms that our measures of semantic variation identify meaning variants of {\em common} words (such as `box' and `believers') that arise in communities of practice. 

% Spe: broadening and narrowing
From qualitative analysis, we observe that contextual specificity, which is significantly higher in words that exhibit a variant at the domain level, can give rise to different semantic phenomena. For instance, in the case of `box'  (see footnote \ref{ftn:box}), we observe semantic \textit{broadening}, a generalisation of meaning possibly as a consequence of metaphorical use.  While in other cases, specificity is related to semantic \textit{narrowing}. This holds, for instance, for `yellow', which has come to mean `yellow card' in the Football domain. The strength of the collocation `yellow card' seems to have made possible a narrower interpretation of `yellow', as in `Terry got a very stupid \textit{yellow}'. 

In contrast to domain-level variation, specificity and frequency do not play an important role across the board for semantic shift at the community level (see Table~\ref{tab:factors}). Meaning variants that are specific to a particular community are not highly frequent and thus it is less likely that they take part in collocations (see e.g., \cite{shin2008beyond}). As mentioned, we find that words with high \textbf{csi} values are less disseminated within the community. We see this as potentially related to the general process of linguistic innovation and diffusion descibed in \cite{chambers1998dialectology} and usually represented by a sigmoid function (see, for example, \cite{fagyal2010centers}). Linguistic variants originate among and are initially adopted by a circumscribed number of members. At this stage (corresponding to the left tail of the function) few users use the innovation, which is therefore not highly disseminated in the community. Our intuition is that the \textbf{csi} index captures innovations which are in this phase. Some variants may then rapidly spread within the community (central part of the function) and possibly to other domain-related communities, until they reach a plateau, in terms of frequency of use (right tail of the function). This is the stage which is captured by our \textbf{dsi} index: the innovation, at this point, has been largely adopted, and, consequently, has a high dissemination value.

It is also possible that some community-specific semantic variants are used as identity markers (e.g., `army' in \re{red.devils} or `believers' in \re{liverpool}), which are then presumably not likely to spread to other communities. Such uses may be limited to members who are particularly invested in the community and thus not part of other domain-related communities, which may lead to lower dissemination (since different communities within a domain share a substantial number of members, as shown in Figure~\ref{fig:reddit}). These speculations, however,  need to be verified with further analysis, which we leave to future work.

\section{Conclusions}
\label{sec:concl}
We have investigated meaning variation from the perspective of social engagement in online communities of practice, exploring the hypothesis that meaning conventions are not only topic dependent, but that different meanings can emerge in communities discussing the same topic. We verified our research hypothesis using a large dataset from Reddit discussion forums, and showed that our quantitative measures allow us to identify semantic variation in the use of common (non-slang) words at both domain and community levels. We evaluated our findings using an extrinsic language modelling task. 

Our analysis of the factors that influence socially-driven semantic variation should be seen as a preliminary investigation, which we believe opens the door to more in-depth studies we plan to conduct in the future. The most natural extension of the current work is an investigation of the social dynamics that lead to meaning variation: while in the present work we have shown the outcome of such dynamics, i.e. the observable meaning shift in different communities of practice, in our future work we plan to focus on the interactions among speakers, which are at the base of observable variation. Directly related to this is the consideration of the diachronic dimension, linking the presence of semantic variation to the more general dynamics of meaning change. 
Our aim in this direction is to consider the evolution of meaning conventions in time while taking into account the network structure of communities of practice.

In parallel, we plan to explore other aspects within the synchronic perspective, such as  the relationship between semantic variation and demographic factors, e.g., geo-location, age, or gender --- in particular, in light of the fact that the datasets we are using here are likely to be biased towards the language use of male speakers. Finally, we want to extend our investigation to a larger set of communities, in order to make our findings and claims more robust.

\bibliographystyle{chicago}
\bibliography{community_lexicon_experiment}

\begin{thebibliography}{}

\bibitem[\protect\citeauthoryear{Altmann, Pierrehumbert, and Motter}{Altmann
  et~al.}{2011}]{altman-etal:plos-one:2011}
Altmann, E.~G., J.~B. Pierrehumbert, and A.~E. Motter (2011).
\newblock Niche as a determinant of word fate in online groups.
\newblock {\em PLoS ONE\/}~{\em 6\/}(5), e19009.

\bibitem[\protect\citeauthoryear{Bamman, Dyer, and Smith}{Bamman
  et~al.}{2014}]{bamman-dyer-smith:2014}
Bamman, D., C.~Dyer, and N.~A. Smith (2014, June).
\newblock Distributed representations of geographically situated language.
\newblock In {\em Proceedings of the 52nd Annual Meeting of the Association for
  Computational Linguistics (Volume 2: Short Papers)}, pp.\  828--834.

\bibitem[\protect\citeauthoryear{Burger, Henderson, Kim, and Zarrella}{Burger
  et~al.}{2011}]{burger2011discriminating}
Burger, J.~D., J.~Henderson, G.~Kim, and G.~Zarrella (2011).
\newblock Discriminating gender on twitter.
\newblock In {\em Proceedings of the Conference on Empirical Methods in Natural
  Language Processing}, pp.\  1301--1309. Association for Computational
  Linguistics.

\bibitem[\protect\citeauthoryear{Cassell and Tversky}{Cassell and
  Tversky}{2005}]{cassell2005language}
Cassell, J. and D.~Tversky (2005).
\newblock The language of online intercultural community formation.
\newblock {\em Journal of Computer-Mediated Communication\/}~{\em 10\/}(2),
  00--00.

\bibitem[\protect\citeauthoryear{Chambers and Trudgill}{Chambers and
  Trudgill}{1998}]{chambers1998dialectology}
Chambers, J.~K. and P.~Trudgill (1998).
\newblock {\em Dialectology}.
\newblock Cambridge University Press.

\bibitem[\protect\citeauthoryear{Ciot, Sonderegger, and Ruths}{Ciot
  et~al.}{2013}]{ciot2013gender}
Ciot, M., M.~Sonderegger, and D.~Ruths (2013).
\newblock Gender inference of twitter users in non-english contexts.
\newblock In {\em EMNLP}, pp.\  1136--1145.

\bibitem[\protect\citeauthoryear{Clark}{Clark}{1996}]{Clark96}
Clark, H.~H. (1996).
\newblock {\em Using language}.
\newblock Cambridge University Press.

\bibitem[\protect\citeauthoryear{Danescu-Niculescu-Mizil, Lee, Pang, and
  Kleinberg}{Danescu-Niculescu-Mizil et~al.}{2012}]{danescu2012echoes}
Danescu-Niculescu-Mizil, C., L.~Lee, B.~Pang, and J.~Kleinberg (2012).
\newblock Echoes of power: Language effects and power differences in social
  interaction.
\newblock In {\em Proceedings of the 21st international conference on World
  Wide Web}, pp.\  699--708. ACM.

\bibitem[\protect\citeauthoryear{Danescu-Niculescu-Mizil, West, Jurafsky,
  Leskovec, and Potts}{Danescu-Niculescu-Mizil et~al.}{2013}]{danescu2013no}
Danescu-Niculescu-Mizil, C., R.~West, D.~Jurafsky, J.~Leskovec, and C.~Potts
  (2013).
\newblock No country for old members: User lifecycle and linguistic change in
  online communities.
\newblock In {\em Proceedings of the 22nd international conference on World
  Wide Web}, pp.\  307--318. ACM.

\bibitem[\protect\citeauthoryear{Doyle}{Doyle}{2014}]{doyle2014mapping}
Doyle, G. (2014).
\newblock Mapping dialectal variation by querying social media.
\newblock In {\em EACL}, pp.\  98--106.

\bibitem[\protect\citeauthoryear{Dunning}{Dunning}{1993}]{dunning1993accurate}
Dunning, T. (1993).
\newblock Accurate methods for the statistics of surprise and coincidence.
\newblock {\em Computational Linguistics\/}~{\em 19\/}(1), 61--74.

\bibitem[\protect\citeauthoryear{Eckert and McConnell-Ginet}{Eckert and
  McConnell-Ginet}{1992}]{eckert-mcconnellginet1992}
Eckert, P. and S.~McConnell-Ginet (1992).
\newblock Communities of practice: Where language, gender, and power all live.
\newblock In K.~Hall, M.~Bucholtz, and B.~Moonwomon (Eds.), {\em Locating
  Power, Proceedings of the 1992 Berkeley Women and Language Conference}, pp.\
  89--99.

\bibitem[\protect\citeauthoryear{Eisenstein, O'Connor, Smith, and
  Xing}{Eisenstein et~al.}{2010}]{eisenstein2010latent}
Eisenstein, J., B.~O'Connor, N.~A. Smith, and E.~P. Xing (2010).
\newblock A latent variable model for geographic lexical variation.
\newblock In {\em Proceedings of the 2010 Conference on Empirical Methods in
  Natural Language Processing}, pp.\  1277--1287. Association for Computational
  Linguistics.

\bibitem[\protect\citeauthoryear{Eisenstein, O'Connor, Smith, and
  Xing}{Eisenstein et~al.}{2014}]{eisenstein2014diffusion}
Eisenstein, J., B.~O'Connor, N.~A. Smith, and E.~P. Xing (2014).
\newblock Diffusion of lexical change in social media.
\newblock {\em PloS one\/}~{\em 9\/}(11), e113114.

\bibitem[\protect\citeauthoryear{Fagyal, Swarup, Escobar, Gasser, and
  Lakkaraju}{Fagyal et~al.}{2010}]{fagyal2010centers}
Fagyal, Z., S.~Swarup, A.~M. Escobar, L.~Gasser, and K.~Lakkaraju (2010).
\newblock Centers and peripheries: Network roles in language change.
\newblock {\em Lingua\/}~{\em 120\/}(8), 2061--2079.

\bibitem[\protect\citeauthoryear{Frermann and Lapata}{Frermann and
  Lapata}{2016}]{frermann2016bayesian}
Frermann, L. and M.~Lapata (2016).
\newblock A bayesian model of diachronic meaning change.
\newblock {\em TACL\/}~{\em 4}, 31--45.

\bibitem[\protect\citeauthoryear{Hamilton, Leskovec, and Jurafsky}{Hamilton
  et~al.}{2016}]{hamilton2016diachronic}
Hamilton, W.~L., J.~Leskovec, and D.~Jurafsky (2016).
\newblock Diachronic word embeddings reveal statistical laws of semantic
  change.
\newblock In {\em Proceedings of ACL 2016}.

\bibitem[\protect\citeauthoryear{Hasan}{Hasan}{1989}]{hasan1989}
Hasan, R. (1989).
\newblock Semantic variation and sociolinguistics.
\newblock {\em Australian Journal of Linguistics\/}~{\em 9\/}(2), 221--275.

\bibitem[\protect\citeauthoryear{Hovy}{Hovy}{2015}]{hovy:2015:ACL-IJCNLP}
Hovy, D. (2015).
\newblock Demographic factors improve classification performance.
\newblock In {\em Proceedings of the 53rd Annual Meeting of the Association for
  Computational Linguistics and the 7th International Joint Conference on
  Natural Language Processing (Volume 1: Long Papers)}, pp.\  752--762.

\bibitem[\protect\citeauthoryear{Huffaker, Jorgensen, Iacobelli, Tepper, and
  Cassell}{Huffaker et~al.}{2006}]{huffaker2006computational}
Huffaker, D., J.~Jorgensen, F.~Iacobelli, P.~Tepper, and J.~Cassell (2006).
\newblock Computational measures for language similarity across time in online
  communities.
\newblock In {\em Proceedings of the HLT-NAACL 2006 workshop on analyzing
  conversations in text and speech}, pp.\  15--22. Association for
  Computational Linguistics.

\bibitem[\protect\citeauthoryear{Jatowt and Duh}{Jatowt and
  Duh}{2014}]{jatowt2014framework}
Jatowt, A. and K.~Duh (2014).
\newblock A framework for analyzing semantic change of words across time.
\newblock In {\em Proceedings of the 14th ACM/IEEE-CS Joint Conference on
  Digital Libraries}, pp.\  229--238. IEEE Press.

\bibitem[\protect\citeauthoryear{Kershaw, Rowe, and Stacey}{Kershaw
  et~al.}{2016}]{kershaw2016towards}
Kershaw, D., M.~Rowe, and P.~Stacey (2016).
\newblock Towards modelling language innovation acceptance in online social
  networks.
\newblock In {\em Proceedings of the Ninth ACM International Conference on Web
  Search and Data Mining}, pp.\  553--562. ACM.

\bibitem[\protect\citeauthoryear{Kingma and Ba}{Kingma and
  Ba}{2014}]{kingma2014adam}
Kingma, D. and J.~Ba (2014).
\newblock Adam: A method for stochastic optimization.
\newblock {\em arXiv preprint arXiv:1412.6980\/}.

\bibitem[\protect\citeauthoryear{Kulkarni, Al-Rfou, Perozzi, and
  Skiena}{Kulkarni et~al.}{2015}]{kulkarni2015statistically}
Kulkarni, V., R.~Al-Rfou, B.~Perozzi, and S.~Skiena (2015).
\newblock Statistically significant detection of linguistic change.
\newblock In {\em Proceedings of the 24th International Conference on World
  Wide Web}, pp.\  625--635. ACM.

\bibitem[\protect\citeauthoryear{Kulkarni, Perozzi, and Skiena}{Kulkarni
  et~al.}{2016}]{kulkarni2016freshman}
Kulkarni, V., B.~Perozzi, and S.~Skiena (2016).
\newblock Freshman or fresher? quantifying the geographic variation of language
  in online social media.
\newblock In {\em ICWSM}, pp.\  615--618.

\bibitem[\protect\citeauthoryear{Manning and Sch\"utze}{Manning and
  Sch\"utze}{1999}]{manning1999foundations}
Manning, C.~D. and H.~Sch\"utze (1999).
\newblock {\em Foundations of statistical natural language processing}.
\newblock MIT Press.

\bibitem[\protect\citeauthoryear{Mikolov, Chen, Corrado, and Dean}{Mikolov
  et~al.}{2013}]{mikolov2013efficient}
Mikolov, T., K.~Chen, G.~Corrado, and J.~Dean (2013).
\newblock Efficient estimation of word representations in vector space.
\newblock {\em arXiv preprint arXiv:1301.3781\/}.

\bibitem[\protect\citeauthoryear{Navigli}{Navigli}{2009}]{navigli2009wsd}
Navigli, R. (2009).
\newblock Word sense disambiguation: A survey.
\newblock {\em ACM Computing Surveys\/}~{\em 41\/}(2), 10.

\bibitem[\protect\citeauthoryear{Nguyen and Ros{\'e}}{Nguyen and
  Ros{\'e}}{2011}]{nguyen2011language}
Nguyen, D. and C.~P. Ros{\'e} (2011).
\newblock Language use as a reflection of socialization in online communities.
\newblock In {\em Proceedings of the Workshop on Languages in Social Media},
  pp.\  76--85. Association for Computational Linguistics.

\bibitem[\protect\citeauthoryear{Nguyen, Trieschnigg, Do\u{g}ru\"{o}z, Gravel,
  Theune, Meder, and De~Jong}{Nguyen et~al.}{2014}]{nguyen2014gender}
Nguyen, D., D.~Trieschnigg, A.~S. Do\u{g}ru\"{o}z, R.~Gravel, M.~Theune,
  T.~Meder, and F.~De~Jong (2014).
\newblock Why gender and age prediction from tweets is hard: Lessons from a
  crowdsourcing experiment.
\newblock In {\em Proceedings of COLING 2014, the 25th International Conference
  on Computational Linguistics: Technical Papers}, pp.\  1950--1961.

\bibitem[\protect\citeauthoryear{Nguyen, Gravel, Trieschnigg, and Meder}{Nguyen
  et~al.}{2013}]{nguyen2013old}
Nguyen, D.-P., R.~Gravel, D.~Trieschnigg, and T.~Meder (2013).
\newblock {``How old do you think I am?'' A study of language and age in
  Twitter}.
\newblock In {\em Proceedings of the Seventh International AAAI Conference on
  Weblogs and Social Media}. AAAI Press.

\bibitem[\protect\citeauthoryear{Noble and Fern\'andez}{Noble and
  Fern\'andez}{2015}]{noble-fernandez:2015:CMCL}
Noble, B. and R.~Fern\'andez (2015, June).
\newblock Centre stage: How social network position shapes linguistic
  coordination.
\newblock In {\em Proceedings of the 6th Workshop on Cognitive Modeling and
  Computational Linguistics}, Denver, Colorado, pp.\  29--38. Association for
  Computational Linguistics.

\bibitem[\protect\citeauthoryear{Pagel, Atkinson, and Meade}{Pagel
  et~al.}{2007}]{pagel2007frequency}
Pagel, M., Q.~D. Atkinson, and A.~Meade (2007).
\newblock Frequency of word-use predicts rates of lexical evolution throughout
  indo-european history.
\newblock {\em Nature\/}~{\em 449\/}(7163), 717--720.

\bibitem[\protect\citeauthoryear{Sagi, Kaufmann, and Clark}{Sagi
  et~al.}{2011}]{sagi2011tracing}
Sagi, E., S.~Kaufmann, and B.~Clark (2011).
\newblock Tracing semantic change with latent semantic analysis.
\newblock {\em Current methods in historical semantics\/}, 161--183.

\bibitem[\protect\citeauthoryear{Shin and Nation}{Shin and
  Nation}{2008}]{shin2008beyond}
Shin, D. and P.~Nation (2008).
\newblock Beyond single words: The most frequent collocations in spoken
  english.
\newblock {\em ELT journal\/}~{\em 62\/}(4), 339--348.

\bibitem[\protect\citeauthoryear{Tran and Ostendorf}{Tran and
  Ostendorf}{2016}]{tran2016characterizing}
Tran, T. and M.~Ostendorf (2016).
\newblock Characterizing the language of online communities and its relation to
  community reception.
\newblock {\em arXiv preprint arXiv:1609.04779\/}.

\bibitem[\protect\citeauthoryear{Velardi and Sclano}{Velardi and
  Sclano}{2007}]{velardi2007termextractor}
Velardi, P. and F.~Sclano (2007).
\newblock "termextractor: a web application to learn the common terminology of
  interest groups and research communities".
\newblock In {\em "7th Conference on Terminology and Artificial Intelligence"},
  pp.\  85--94.

\bibitem[\protect\citeauthoryear{Wenger}{Wenger}{2000}]{wenger2000communities}
Wenger, E. (2000).
\newblock {\em Communities of practice: Learning, meaning, and identity}.
\newblock Cambridge University Press.

\bibitem[\protect\citeauthoryear{Wijaya and Yeniterzi}{Wijaya and
  Yeniterzi}{2011}]{wijaya2011understanding}
Wijaya, D.~T. and R.~Yeniterzi (2011).
\newblock Understanding semantic change of words over centuries.
\newblock In {\em Proceedings of the 2011 international workshop on DETecting
  and Exploiting Cultural diversiTy on the social web}, pp.\  35--40. ACM.

\bibitem[\protect\citeauthoryear{Yang and Eisenstein}{Yang and
  Eisenstein}{2017}]{YangEisensteinTACL2017}
Yang, Y. and J.~Eisenstein (2017).
\newblock Overcoming language variation in sentiment analysis with social
  attention.
\newblock {\em Transactions of the Association for Computational
  Linguistics\/}.

\bibitem[\protect\citeauthoryear{Yarowsky}{Yarowsky}{2010}]{yarowsky2010wsd}
Yarowsky, D. (2010).
\newblock Word sense disambiguation.
\newblock In {\em Handbook of Natural Language Processing, Second Edition},
  pp.\  315--338. Chapman and Hall/CRC.

\bibitem[\protect\citeauthoryear{Zaremba, Sutskever, and Vinyals}{Zaremba
  et~al.}{2014}]{zaremba2014recurrent}
Zaremba, W., I.~Sutskever, and O.~Vinyals (2014).
\newblock Recurrent neural network regularization.
\newblock {\em arXiv preprint arXiv:1409.2329\/}.

\bibitem[\protect\citeauthoryear{Zhang, Jatowt, Bhowmick, and Tanaka}{Zhang
  et~al.}{2015}]{zhang2015omnia}
Zhang, Y., A.~Jatowt, S.~S. Bhowmick, and K.~Tanaka (2015).
\newblock Omnia mutantur, nihil interit: Connecting past with present by
  find-ing corresponding terms across time.
\newblock In {\em Proc. of ACL}, pp.\  645--655.

\bibitem[\protect\citeauthoryear{Zipf}{Zipf}{1949}]{Zipf49}
Zipf, G. (1949).
\newblock {\em Human Behavior and the Principle of Least Effort: an
  Introduction To Human Ecology}.
\newblock Addison-Wesley.

\end{thebibliography}
\end{document}